\documentclass[10pt,twocolumn,letterpaper]{article}

\usepackage{cvpr}              

\definecolor{cvprblue}{rgb}{0.21,0.49,0.74}
\usepackage[pagebackref,breaklinks,colorlinks,allcolors=cvprblue]{hyperref}
\usepackage[accsupp]{axessibility}  
\usepackage{multirow}
\def\paperID{MAI-3} 
\def\confName{CVPR}
\def\confYear{2025}

\title{Robust 6DoF Pose Estimation Against Depth Noise and a Comprehensive Evaluation on a Mobile Dataset}

\author{
  \textbf{Zixun Huang$^1$}\footnotemark[1]
  ~~~
  \textbf{Keling Yao$^2$}\footnotemark[1]
  ~~~
  \textbf{Seth Z. Zhao$^3$}
\\
  \textbf{Chuanyu Pan$^1$} 
  ~~~
  \textbf{Allen Y. Yang$^1$}\footnotemark[2] 
  \\
  $^1$UC Berkeley~~~~~~$^2$CMU~~~~~~$^3$UCLA
}

\begin{document}
\maketitle

\begin{center}
    \renewcommand{\thefootnote}{\fnsymbol{footnote}}
    \footnotetext[1]{Equal contributions. Contact: \texttt{zixun@berkeley.edu}.}
    \footnotetext[2]{Corresponding Author. Contact: \texttt{yang@eecs.berkeley.edu}}
    \renewcommand{\thefootnote}{\arabic{footnote}}
\end{center}

\def\thefootnote{\arabic{footnote}}
\begin{abstract}
Robust 6DoF pose estimation with mobile devices is the foundation for applications in robotics, augmented reality, and digital twin localization. In this paper, we extensively investigate the robustness of existing RGBD-based 6DoF pose estimation methods against varying levels of depth sensor noise. We highlight that existing 6DoF pose estimation methods suffer significant performance discrepancies due to depth measurement inaccuracies. In response to the robustness issue, we present a simple and effective transformer-based 6DoF pose estimation approach called DTTDNet\footnote{This work was previously presented as a non-archival poster at the 2024 ICML DMLR Workshop. This submission does not overlap with any archival publications.}, featuring a novel geometric feature filtering module and a Chamfer distance loss for training. Moreover, we advance the field of robust 6DoF pose estimation and introduce a new dataset -- Digital Twin Tracking Dataset Mobile (DTTD-Mobile), tailored for digital twin object tracking with noisy depth data from the mobile RGBD sensor suite of the Apple iPhone 14 Pro. Extensive experiments demonstrate that DTTDNet significantly outperforms state-of-the-art methods at least 4.32, up to 60.74 points in ADD metrics on the DTTD-Mobile. More importantly, our approach exhibits superior robustness to varying levels of measurement noise, setting a new benchmark for robustness to measurement noise. The project page is publicly available at \href{https://openark-berkeley.github.io/DTTDNet/}{https://openark-berkeley.github.io/DTTDNet/}.
\end{abstract}    
\section{Introduction}
\label{sec:intro}
Six-degrees-of-freedom (6DoF) object pose estimation aims at determining the position and orientation of an object in 3D space. In contrast to the more matured technology of camera tracking in static settings known as visual odometry or simultaneous localization and mapping \cite{ORBSLAM3_TRO,8954208,applearkit,intelrealsense}, identifying the relative position and orientation of one or more objects with respect to the user's ego position is a core function essential for ensuring a high-quality user experience in applications like augmented reality (AR). In the most general setting, each object with respect to the ego position may undergo independent rigid-body motion, and the combined effect of overlaying multiple objects in the scene may also cause parts of the objects to be occluded from the measurement of the ego position. In this paper, the main topic of our investigation is to study the 6DoF pose estimation problem under a wide range of motion, occlusion, color, and lighting conditions, especially improving the accuracy and robustness of algorithms under novel data sensor properties. The dataset and proposed model are made publicly available at \href{https://github.com/augcog/DTTD2}{https://github.com/augcog/DTTD2}.

\begin{figure}
        \centering
\includegraphics[width=1.0\linewidth]{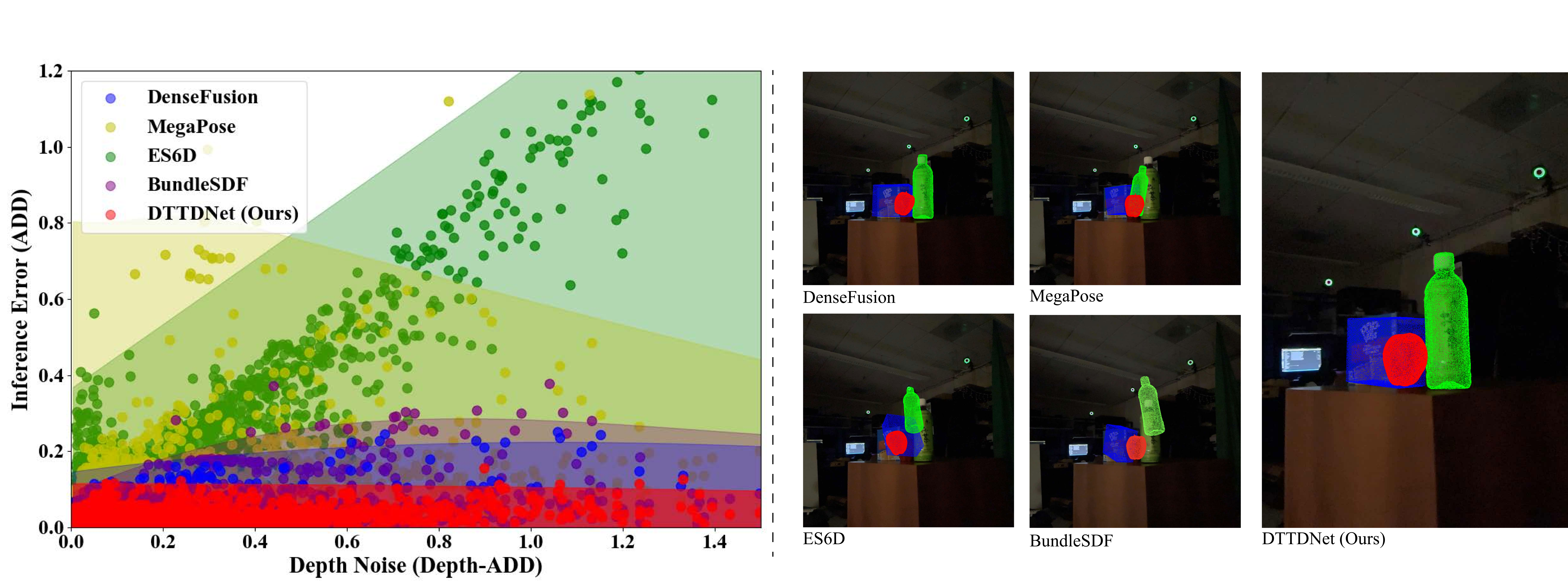}
    \caption{\textit{Left}: Shadow plot of the relation between the \textbf{depth noise} (\emph{depth-ADD}) and the \textbf{inference error} (ADD) of considered state-of-the-art methods and proposed DTTDNet.
    \textit{Right}: Visualization of pose estimation results of baseline methods and proposed DTTDNet.} 
\label{fig:hero_img}
\vspace{-12pt}
\end{figure}

Recent advancements in the field of 6DoF pose estimation have primarily been motivated by deep neural network (DNN) approaches that advocate end-to-end training to carry out crucial tasks such as image semantic segmentation, object classification, and object pose estimation. Notable studies \cite{wang2019densefusion, He_2020_CVPR, He_2021_CVPR, Jiang_2022_CVPR, mo2022es6d} have demonstrated the effectiveness of these pose estimation algorithms using established real-world 6DoF pose estimation datasets \cite{10.1007/978-3-642-37331-2_42, marion2018label, xiang2018posecnn, liu2020keypose, hodan2017tless, liu2021stereobj1m}. However, it should be noted that these datasets primarily focus on robotic grasping tasks, and applying these solutions to environments served with mobile devices introduces a fresh set of challenges. A previous work \cite{DTTD} first studied this gap in the context of 6DoF pose estimation and replicated real-world digital-twin scenarios with varying levels of capture distances, lighting conditions, and object occlusions. It is important to mention that, this dataset was collected using Microsoft Azure Kinect, which may not be the most suitable camera platform for studying 3D localization under realistic mobile environments.

Alternatively, Apple has emerged as a strong proponent of utilizing RGB-D spatial sensors for mobile AR applications with the design of their iPhone Pro camera suite, such as on the latest Apple iPhone 14 Pro model. This particular smartphone is equipped with a rear-facing LiDAR depth sensor \cite{ilci2020high, li2016vehicle, bu2019pedestrian, gu2021ecpc, you2019pseudo, weng2019monocular}, a critical component to achieving accurate and detailed 3D perception and spatial understanding. However, one distinguishing drawback of the iPhone LiDAR depth is the low resolution of the depth map produced by the iPhone ARKit \cite{applearkit}, a $256\times 192$ resolution compared to a $1280\times 720$ depth map provided by the Microsoft Azure Kinect. This low resolution is exacerbated by large errors in the retrieved depth map, which is also observed in \cite{iphonesensor1, iphonesensor2}. The large amounts of errors (as shown in Fig.~\ref{fig:objects_lidar}) in the iPhone data also pose challenges for researchers to develop a pose estimator that can correctly predict object poses that rely heavily on the observed depth map, which has not been particularly addressed in previous works \cite{wang2019densefusion, sun2022onepose, gu2021ecpc,you2019pseudo, weng2019monocular, mo2022es6d}. We will demonstrate this in following experiments (Table~\ref{tab:eval_dttb_ycb}).

To investigate the 6DoF pose estimation problem under the most popular mobile depth sensor, namely, the Apple iPhone 14 Pro LiDAR, we propose an RGBD-based transformer model for 6DoF object pose estimation, which is designed to effectively handle inaccurate depth measurements and noise. As shown in Fig. \ref{fig:hero_img}, our method shows robustness against noisy depth input, while other baselines failed in such conditions. Meanwhile, we introduce DTTD-Mobile, a novel RGB-D dataset captured by iPhone 14 Pro, to bridge the gap of digital-twin pose estimation with mobile devices, allowing research into extending algorithms to iPhone data and analyzing the unique nature of iPhone depth sensors. Our contributions are summarized into three parts:
\begin{enumerate}
    \item We propose a new transformer-based 6DoF pose estimator with depth-robust designs on modality fusion and training strategies, called DTTDNet. The new solution outperforms other state-of-the-art methods by a large margin especially in noisy depth conditions. 
    \item We introduce DTTD-Mobile as a novel digital-twin pose estimation dataset captured with mobile devices. We provide in-depth LiDAR depth analysis and evaluation metrics to illustrate the unique properties and complexities of mobile LiDAR data.
    \item We conduct extensive experiments and ablation studies to demonstrate the efficacy of DTTDNet and shed light on how the depth robustifying module works.
\end{enumerate}

\begin{figure}
    \centering\includegraphics[width=1.0\linewidth]{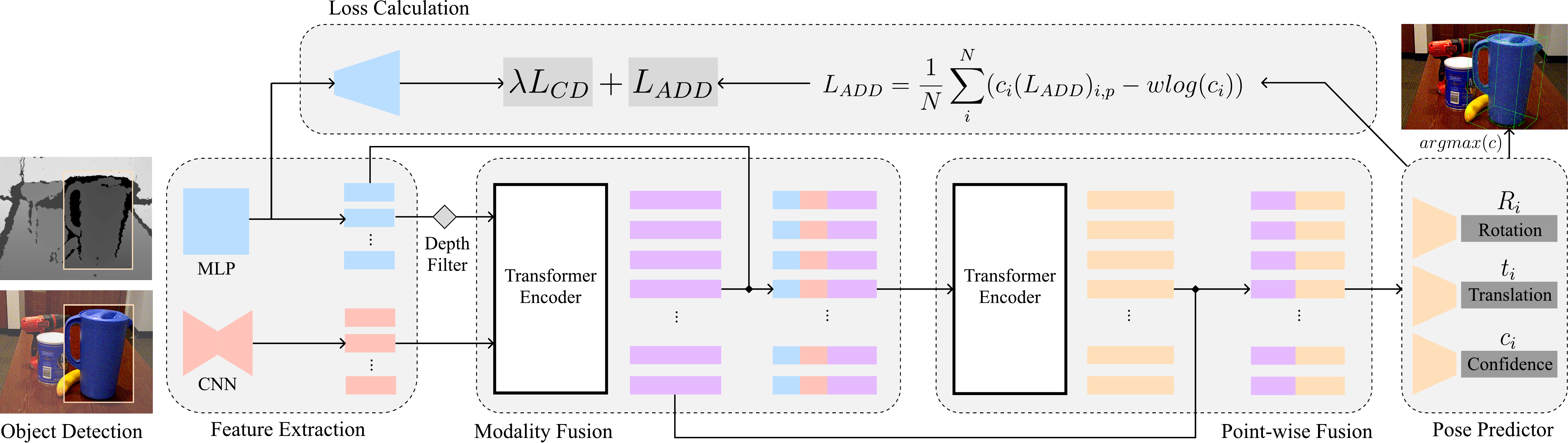}
    \caption{\textbf{Model Architecture Overview.} DTTDNet pipeline starts with segmented depth maps and cropped RGB images. The point cloud from the depth map and RGB colors are encoded and integrated point-wise. Extracted features are then fed into an attention-based two-stage fusion. Finally, the pose predictor produces point-wise predictions with both rotation and translation. }
    \label{fig:modelArch}
\end{figure}

\section{Methods}
In this section, we elaborate on the details of our method. The objective is to estimate the 3D location and pose of a known object in the camera coordinates from the RGBD images. This position can be represented using homogeneous transformation matrix $p\in SE(3)$, which consists of a rotation matrix $R\in SO(3)$ and a translation matrix $t\in\mathbb{R}^3$, $p=[R|t]$. Section \ref{architecture} describes our transformer-based model architecture. Section \ref{depth_data_robust} introduces two depth robustifying modules on depth feature extractions, dedicated to geometric feature reconstruction and filtering. Section \ref{attention-based-fusion} illustrates our modality fusion design for the model to disregard significant noisy depth features. Finally, Section \ref{learning_objective} describes our final learning objective.

\subsection{Architecture Overview}
\label{architecture}
Fig. \ref{fig:modelArch} illustrates the overall architecture of the proposed DTTDNet. The DTTDNet pipeline takes segmented depth maps and cropped RGB images as input. It then obtains feature embedding for both RGB and depth images through separate CNN and point-cloud encoders on cropped RGB images and reconstructed point cloud corresponding to the cropped depth images.\footnote{We preprocessed the RGB and depth images to guarantee the pixel-level correspondence between the RGB image and the depth image. The preprocessing process is detailed in Section \ref{sec:Dataset}.} Inspired by PSPNet \cite{zhao2017pyramid}, the image embedding network comprises a ResNet-18 encoder, which is then followed by 4 up-sampling layers acting as the decoder. It translates an image of size $H\times W\times 3$ into a $H\times W\times d_{rgb}$ embedding space. For depth feature extraction, we take segmented depth pixels and transform them into 3D point clouds with the camera intrinsic. 

The 3D point clouds are initially processed using an auto-encoder inspired by the PointNet \cite{qi2017pointnet}. The PointNet-style encoding step aims to capture geometric representations in latent space in $\mathbb{R}^{d_{1}}$. In this context, the encoder component produces two sets of features: early-stage point-wise features in $\mathbb{R}^{N\times d_{2}}$ and global geometric features in $\mathbb{R}^{d_{3}}$. Subsequently, we add a decoder that is guided by a reference point set $P$ to generate the predicted point cloud $\hat{P}$. Features extracted from the encoder are subsequently combined with the learned representations to create a new feature sequence with a dimension of $\mathbb{R}^{N\times d_{geo}}$, where $d_{geo} = d_{1} + d_{2} + d_{3}$.

This results in a sequence of geometric tokens with a length equal to the number of points $N$. Extracted RGB and depth features are then fed into a two-stage attention-based fusion block, which consists of modality fusion and point-wise fusion. 
Finally, the pose predictor produces point-wise predictions with both rotation and translation. The predictions are then voted based on unsupervised confidence scoring to get the final 6DoF pose estimate.

\subsection{Design for Robustifying Depth Data}
\label{depth_data_robust}

\begin{figure}
    \includegraphics[width=1.0\linewidth]{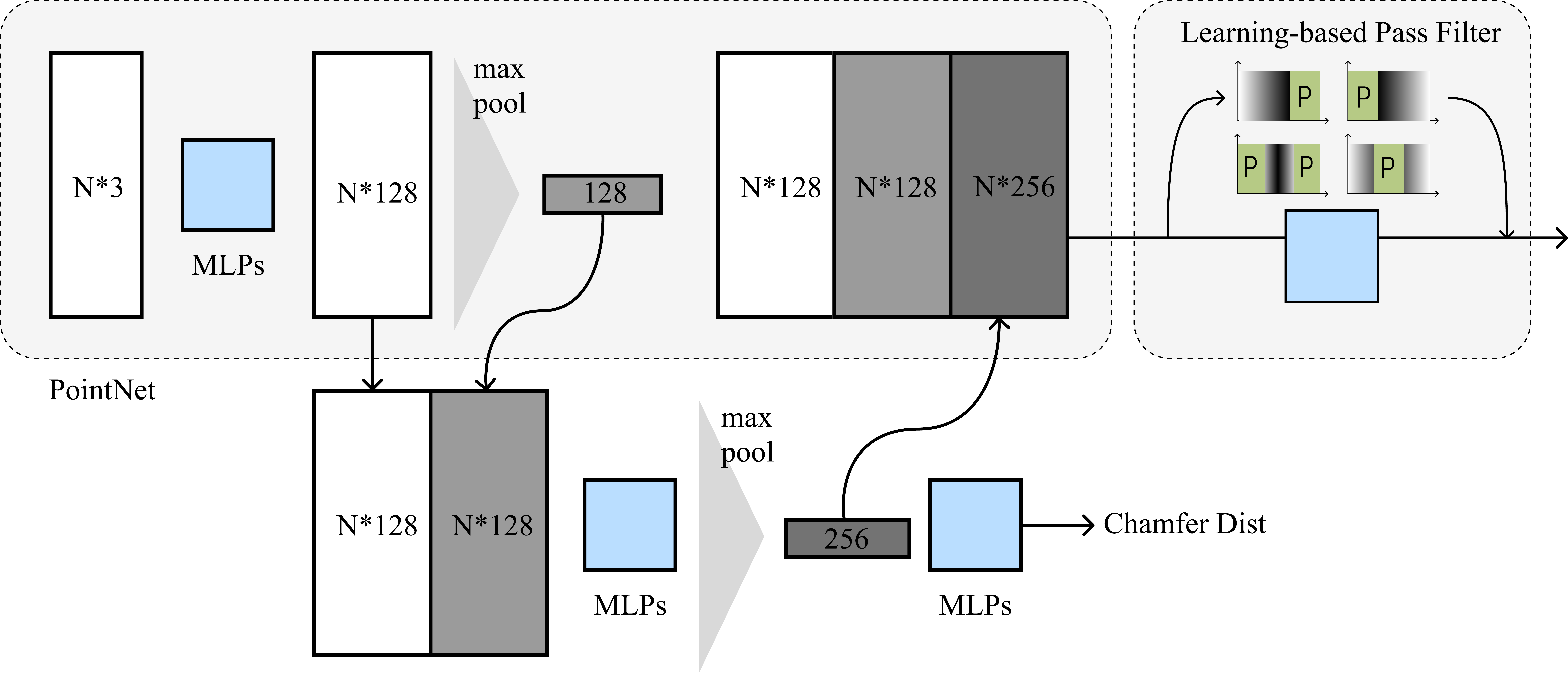}
    \caption{{Chamfer distance loss and geometric feature filtering.} }
    \label{fig:cdlgff}
\end{figure}

In this section, we will introduce two modules (Fig. \ref{fig:cdlgff}) that enable the point-cloud encoder in DTTDNet to handle noisy and low-resolution LiDAR data robustly.

\noindent \textbf{Chamfer Distance Loss.} \label{cdl}Past methods either treated the depth information directly as image channels \cite{mo2022es6d} or directly extracted features from a point cloud for information extraction \cite{wang2019densefusion}. These methods underestimated the corruption of the depth data caused by noise and error during the data collection process. To address this, we first introduce a downstream task for point-cloud reconstruction and utilize the Chamfer distance as a loss function to assist our feature embedding in filtering out noise. The Chamfer distance loss is widely used for denoising in 3D point clouds \cite{hermosilla2019total, duan20193d}, and it is defined as the following equation between two point clouds $P\in\mathbb{R}^{N\times3}$ and $\hat{P}\in\mathbb{R}^{N\times3}$\label{cdl_eqa}:
\begin{equation}
    \resizebox{.95\hsize}{!}{$L_{CD}(\hat{P},P)=\frac{1}{N}(\sum\limits_{\hat{x_i}\in\hat{P}}\min\limits_{x_j\in P}\left \| x_i-\hat{x_j} \right \|_2^2+\sum\limits_{x_i\in P}\min\limits_{\hat{x_j}\in\hat{P}}\left \| x_i-\hat{x_j} \right \|_2^2)$}
\end{equation}
where $\hat{P}$ denotes the decoded point set from the embedding, and $P$ denotes the reference point set employed to guide the decoder's learning. For the reference point set, we use the point cloud sampled from the corresponding object CAD models, which are used only in the training process.


\noindent \textbf{Geometric Feature Filtering.} Due to the non-Gaussian noise distribution in iPhone LiDAR data (Fig. \ref{fig:objects_lidar}), which should be assumed for most depth camera data, normal estimators might either get perturbed by such noisy features or interpret wrong camera-object rotations. To deal with this sensor-level error, we advocate for the integration of a geometric feature filtering (GFF) module before the modality fusion module. Our approach incorporates the fast Fourier transform (FFT) into the geometric feature encoding. Specifically, the GFF module includes an FFT, a subsequent single layer of MLP, and finally, an inverse-FFT. By leveraging FFT, we can transpose the input sequence of geometric signals to the frequency domain, which selects significant features from noisy input signals. After that, we obtain a more refined geometric embedding that is resilient to the non-Gaussian iPhone LiDAR noise.

\subsection{Attention-based RGBD Fusion}
\label{attention-based-fusion}
Previous papers have emphasized the importance of modality fusion \cite{He_2021_CVPR, wang2019densefusion} and the benefits of gathering nearest points from the point cloud \cite{mo2022es6d, He_2021_CVPR} in RGBD-based pose estimation tasks. While the feature extractor widens each point's receptive field, we aim for features to interact beyond their corresponding points \cite{wang2019densefusion} or neighboring points \cite{He_2021_CVPR}. In predicting the 6DoF pose of a cuboid based on multiple feature descriptors, our focus is on attending to various corner points, rather than solely those in close proximity to each other. To this end, inspired by recent transformer-based models used for modality fusion \cite{dosovitskiy2021image, he2021masked, paul2021vision, li2023token, radford2021learning, kim2021vilt, SMD, cl_speech}, we leverage the self-attention mechanism \cite{vaswani2023attention} to amplify and integrate important features while disregarding the significant LiDAR noise. Specifically, our fusion part is divided into two stages: modality fusion and point-wise fusion (Fig. \ref{fig:modelArch}). Both of our fusion modules consist of a standard transformer encoder with linear projection, multi-head attention and layer norm. The former module utilizes the embedding from single-modal encoders and feeds them into a transformer encoder in parallel for cross-modal fusion. The latter fusion module relies on similarity scores among points. It merges all feature embedding in a point-wise manner before feeding them into a transformer encoder. Detailed design and visual analysis of 2 fusion stages are described in our supplemental materials.

\subsection{Learning Objective}
\label{learning_objective}
Based on the overall network structure, our learning objective is to perform 6DoF pose regression, which measures the disparity between points sampled on the object's model in its ground truth pose and corresponding points on the same model transformed by the predicted pose. Specifically, the pose estimation loss is defined as:
\begin{equation}
    (L_{ADD})_{i,p} = \frac{1}{m}\sum_{x \in M} \| (Rx + t) - (\hat{R_i}x + \hat{t_i})\|
\end{equation}
where $M\in \mathbb{R}^{m\times3}$ represents the randomly sampled point set from the object's 3D model, $p=[R|t]$ denotes the ground truth pose, and $\hat{p_i}=[\hat{R_i}|\hat{t_i}]$ denotes the predicted pose generated from the fused feature of the $i^{th}$ point. Our objective is to minimize the sum of the losses for each fusion point, which can be expressed as $L_{ADD}=\frac{1}{N}\sum_{i}^N(L_{ADD})_{i,p}$, where $N$ is the number of randomly sampled points (token sequence length in the point-wise fusion stage). 
Meanwhile, we introduce a confidence regularization score ($c_i$) along with each prediction $\hat{p_i}=[\hat{R_i}|\hat{t_i}]$, which denotes confidence among the predictions for each fusion point:
\begin{equation}
    L_{ADD} = \frac{1}{N}\sum_{i}^N (c_i(L_{ADD})_{i,p} - w log(c_i))
\end{equation}

Predictions with low confidence will lead to a low ADD loss, but this will be balanced by a high penalty from the second term with hyper-parameter $w$. 
Finally, the Chamfer distance loss, as outlined in Section \ref{cdl_eqa}, undergoes joint training throughout the training process, leading us to derive our ultimate learning objective as follows:
\begin{equation}
    L = L_{ADD}+\lambda L_{CD}
\end{equation}
where $\lambda$ denotes the weight of the Chamfer distance loss.

\begin{table*}
    \centering
    \caption{Features and statistics of different datasets.}
    \label{tab:dataset_comparison}
    \resizebox{0.9 \textwidth}{!}{
    \begin{tabular}{l|c|c|c|c|c|c|c|c|c}
    \toprule
        Dataset & Modality & iPhone Camera & Texture & Occlusion  & Light variation & \# of frames & \# of scenes & \# of objects & \# of annotations\\
    \midrule
        \text{StereoOBJ-1M \cite{liu2021stereobj1m}} & RGB & $\times$ & $\checkmark$ &  $\checkmark$  & $\checkmark$ & 393,612 & 182 & 18 & 1,508,327 \\
        \text{LINEMOD \cite{10.1007/978-3-642-37331-2_42}} & RGBD & $\times$ & $\checkmark$ & $\checkmark$ & $\times$ & 18,000 & 15 & 15 & 15,784  \\
        \text{YCB-Video \cite{xiang2018posecnn}}& RGBD  & $\times$ & $\checkmark$ & $\checkmark$ & $\times$ & 133,936 & 92 & 21 & 613,917  \\
        \text{DTTD \cite{DTTD}} & RGBD  & $\times$ & $\checkmark$ & $\checkmark$  & $\checkmark$ & 55,691 & 103 & 10 & 136,226 \\
        \text{TOD \cite{liu2020keypose}} & RGBD  & $\times$ & $\checkmark$ & $\times$ & $\times$ & 64,000 & 10 & 20 & 64,000  \\
        \text{LabelFusion \cite{marion2018label}} & RGBD  & $\times$ & $\checkmark$ &  $\checkmark$  & $\checkmark$ & 352,000 & 138 & 12 & 1,000,000  \\
        \text{T-LESS \cite{hodan2017tless}} & RGBD & $\times$ & $\times$ & $\checkmark$ & $\times$ & 47,762 & - & 30 & 47,762  \\
        \textbf{DTTD-Mobile (Ours)}& RGBD & $\checkmark$ & $\checkmark$ & $\checkmark$  & $\checkmark$ & 47,668 & 100 & 18 & 114,143 \\
    \bottomrule
    \end{tabular}}
\end{table*}

\section{Dataset Description} \label{sec:Dataset}
\begin{figure*}[tb]
\vspace{-6pt}
\centering
\includegraphics[width=0.9 \linewidth]{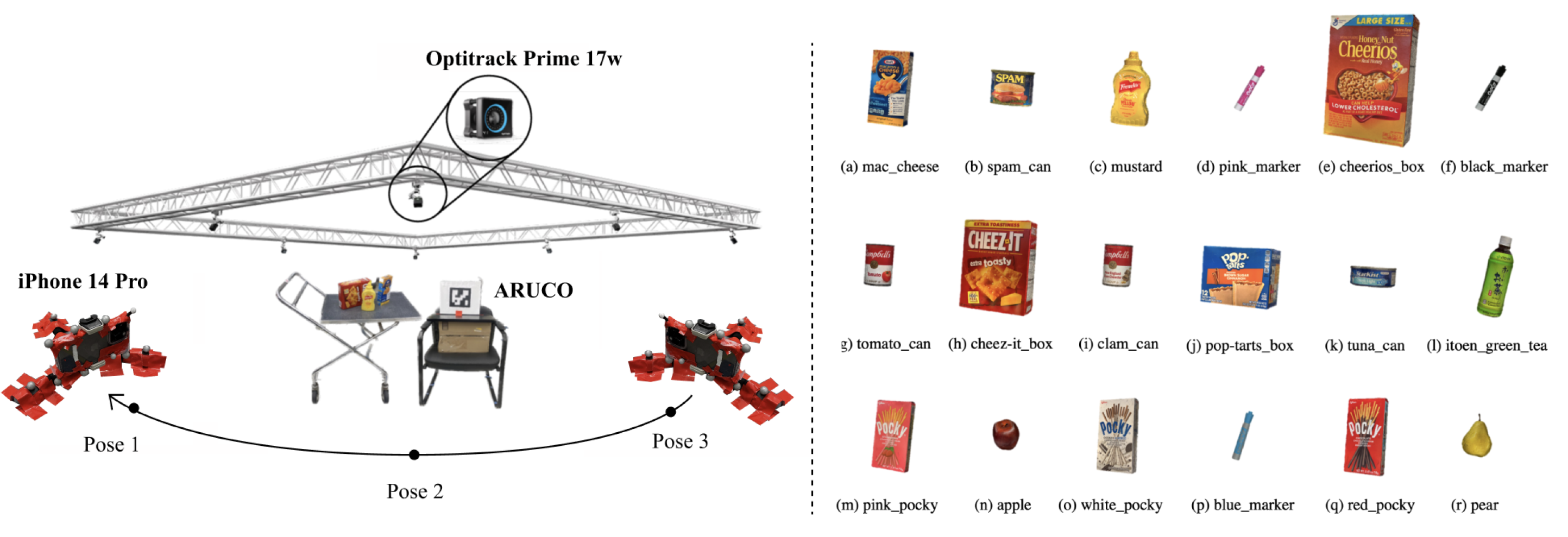}
\caption{\textit{Left:} Setup of our data acquisition pipeline. \textit{Right:} 3D models of the 18 objects in DTTD-Mobile.}
\label{fig:objects}
\vspace{-12pt}
\end{figure*}

DTTD-Mobile dataset contains 18 rigid objects along with their textured 3D models. The data are generated from 100 scenes, each of which features one or more of the objects in various orientations and occlusion. Following \cite{DTTD}, our data generation pipeline is consists of using a professional OptiTrack motion capture system that captures camera pose along the scene and using Apple's ARKit\footnote{https://developer.apple.com/documentation/arkit/} framework to capture RGB images from the iPhone camera and depth information from LiDAR scanner, as illustrated in Fig. \ref{fig:objects}. After obtaining such data, we then use the open-sourced data annotation pipeline provided by \cite{DTTD} to annotate ground-truth object poses. Through this pipeline, the dataset offers ground-truth labels for 3D object poses and per-pixel semantic segmentation. Additionally, it provides detailed camera specifications, pinhole camera projection matrices, and distortion coefficients. Detailed features and statistics are presented in Table \ref{tab:dataset_comparison}. The fact that the DTTD-Mobile dataset includes multiple sets of geometrically similar objects, each having distinct color textures, poses challenges to existing digital-twin localization solutions.  To ensure compatibility with other existing datasets, some of the collected objects partially overlap with the YCB-Video \cite{xiang2018posecnn} and DTTD \cite{DTTD} datasets. Specific details on data acquisition, benchmarking, and evaluation are provided in the appendix.

\begin{figure*}
\includegraphics[width=1.0 \linewidth]{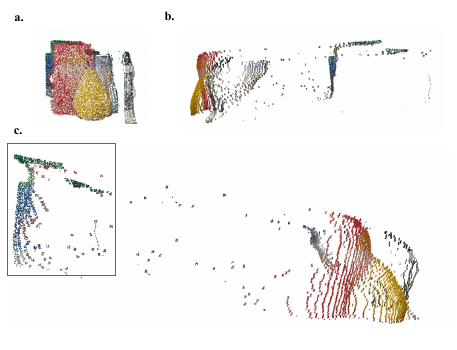}
\caption{Visualization of an iPhone LiDAR depth scene that shows distortion and long-tail non-Gaussian noise (highlighted inside the red box). (a) Front view. (b) Left view. (c) Right view.}
\label{fig:objects_lidar}
\end{figure*}

\section{iPhone LiDAR data analysis}\label{depthADD}
Compared to dedicated depth cameras such as the Microsoft Azure Kinect or Intel Realsense, iPhone 14 Pro LiDAR exhibits more noise and lower resolution at $256 \times 192$ depth maps, which leads to high magnitudes of distortion on objects' surfaces. Additionally, it introduces long-tail noise on the projection edges of objects when performing interpolation operations between RGB and depth features. Fig. \ref{fig:objects_lidar} demonstrates one such example of iPhone 14 Pro's noisy depth data.

To further quantitatively assess the depth noise of each object from the iPhone's LiDAR, we analyze the numerical difference between \emph{LiDAR-measured depth map}, which is acquired directly from iPhone LiDAR, and \emph{reference depth map}, which is derived through ground truth pose annotations. Specifically, to obtain the reference depth map, we leverage ground truth annotated object poses to render the depth projections of each object. We then apply the segmentation mask associated with each object to filter out depth readings that might be compromised due to occlusion. To measure the difference between ground truth and reference depth map, we introduce the \emph{depth-ADD} metric, which calculates the average pixel-wise L1 distance between the ground truth depth map and the reference depth map in each frame. The \emph{depth-ADD} value of each object at frame $n$ is calculated as follows: 
\begin{equation}
    \mathrm{depth{-}ADD}_n = \frac{1}{d}\sum_{i \in D} \left | \mathrm{depth_{LiDAR}}_i - \mathrm{depth_{ref}}_i \right |,
\end{equation}
where $D$ denotes the LiDAR depth map and $i$ denotes the index of pixels on it. $\mathrm{depth_{LiDAR}}_i$ and $\mathrm{depth_{ref}}_i$ represent the depth values from $D$ and the corresponding depth value from the reference depth map. The set $D$ encompasses all indices $i$ under an object's segmentation mask where both $\mathrm{depth_{LiDAR}}_i$ and $\mathrm{depth_{ref}}_i$ yield values greater than zero. The final \emph{depth-ADD} value of each object is the average of such measurements across all $N$ frames:
\begin{equation}
    \mathrm{depth{-}ADD} = \frac{1}{N}\sum_{n \in N}\mathrm{depth{-}ADD}_n
\end{equation}
Tab. \ref{tab:eval_dttb} includes the average \emph{depth-ADD} error in each sampled object in the second column. Greater \emph{depth-ADD} values indicate increased distortions and the presence of long-tail noise in the depth data. Our analysis indicates that the mean \emph{depth-ADD} across all objects is around 0.25m. It is worth noticing that the depth quality varies significantly and could be affected by outliers. For example, there are three objects: \textit{black\_marker}, \textit{blue\_marker} and \textit{pink\_marker} exhibiting greater errors in comparison with the other objects. Detailed depth analysis is reported in the appendix.

\section{Experiments}
\begin{table*}
    \centering
    \caption{\textbf{Comparison with diverse 6DoF pose estimation baselines on DTTD-Mobile dataset.} We showcase AUC results of ADD-S and ADD on all 18 objects, higher is better. Based on considered 4 baselines, our model significantly improves the accuracy on most objects. Note that the left-most column indicates the per-object \emph{depth-ADD} error. }
    \label{tab:eval_dttb}
    \resizebox{1.0 \textwidth}{!}{
    \begin{tabular}{|l|c|c|c|c|c|c|c|c|c|c|c|}
    \toprule
        & \emph{depth-ADD}
        & \multicolumn{2}{|c|}{DenseFusion \cite{wang2019densefusion}} 
         & \multicolumn{2}{|c|}{MegaPose-RGBD \cite{labbé2022megapose}}  
         & \multicolumn{2}{|c|}{ES6D \cite{mo2022es6d}} 
         & \multicolumn{2}{|c|}{BundleSDF \cite{wen2023bundlesdf}} 
         & \multicolumn{2}{|c|}{DTTDNet (Ours)}
\\
    \midrule
         Object & 
         Average &
         \multicolumn{1}{|c|}{ADD AUC} &
         \multicolumn{1}{|c|}{ADD-S AUC} &
         \multicolumn{1}{|c|}{ADD AUC} &
         \multicolumn{1}{|c|}{ADD-S AUC} &
         \multicolumn{1}{|c|}{ADD AUC}  &
         \multicolumn{1}{|c|}{ADD-S AUC}  & 
         \multicolumn{1}{|c|}{ADD AUC}  &
         \multicolumn{1}{|c|}{ADD-S AUC}  &
         \multicolumn{1}{|c|}{ADD AUC}  &
         \multicolumn{1}{|c|}{ADD-S AUC} 
\\
    \midrule
         \text{mac\_cheese}& 0.184  & 88.10 & 93.17 &78.98&87.94& 28.29 &   57.06& 89.95& 94.84& \textbf{94.06}&\textbf{97.02}\\
         
         \text{tomato\_can}  & 0.222 & 69.10 &  93.42&68.85&84.48& 19.07  & 56.17 & 79.62& 93.65&\textbf{74.23}&\textbf{94.01}\\
         
         \text{tuna\_can}& 0.278  & 42.90& 79.94 &8.90&22.11&10.74  &  26.86 & 25.05& 37.94&\textbf{62.98}&\textbf{87.05}\\
         
         \text{cereal\_box} & 0.151 & 75.20 & 88.12 &59.89&71.53& 10.09  &53.92 & 0.00& 0.00&\textbf{86.55}&\textbf{92.74}  \\
         
         \text{clam\_can} & 0.157 &\textbf{90.49}  &  96.32&74.11&90.45&17.75  & 35.92 & 75.22& 96.05&88.15&\textbf{96.92}\\
         
         \text{spam} & 0.286 &53.29 & 91.14  &72.35&86.16& 3.17 & 13.74 & \textbf{89.24}& \textbf{95.12}&52.81&90.83\\
         
         \text{cheez-it\_box} & 0.152 &  82.73 & 92.10&\textbf{89.18}&\textbf{94.83}& 7.81 & 37.14 & 42.46& 51.69& 87.03&93.91\\
         
         \text{mustard} & 0.184  &78.41   & 91.31&76.08&85.38& 21.89 & 52.56 & 84.03& \textbf{92.99}&\textbf{84.06}&{92.15} \\
         
         \text{pop-tarts\_box} & 0.139 & 82.94& 92.58  &44.36&58.97& 3.44 & 35.26 & 82.24& 92.01&\textbf{84.55}&\textbf{92.65}\\
         
         \text{black\_marker} & 0.769  & 32.22 & 38.72 &17.38&34.15& 2.12 & 3.72 & 0.00& 0.00&\textbf{44.08}&\textbf{53.50}\\
         
         \text{blue\_marker}  & 0.370 & \textbf{66.06} & \textbf{74.80} &6.87&12.46& 16.88  & 41.46& 0.00 & 0.00& 50.88&61.69\\
         
         \text{pink\_marker} & 0.410 &  56.46&67.86  &47.84&58.59& 1.59 & 7.01 & 0.00 & 0.00 &\textbf{64.18}&\textbf{73.00} \\
         
         \text{green\_tea} & 0.265  & 64.37 & \textbf{93.10} &48.43&70.50&8.80  & 32.86 & 60.24 & 87.29 & \textbf{64.59}&92.31\\
         
         \text{apple} & 0.119 & 68.97 &  91.13&32.85&76.43&  31.65 &58.46 & 79.27 & 91.78 &\textbf{82.45} &\textbf{94.80} \\
         
         \text{pear} & 0.085  & {65.66} & {91.31} &35.80&56.73&  16.93  &32.57& \textbf{80.12} & \textbf{93.72} &47.83&88.11 \\
         
         \text{pink\_pocky}& 0.231  &50.64 &  67.17 &8.69&18.25&  0.77& 1.93 & 2.31 & 6.25 & \textbf{61.40}&\textbf{82.33} \\
         
         \text{red\_pocky} & 0.245  &88.14 & 93.76  &76.49&84.56&   25.32&51.16 & 77.08& 91.26& \textbf{90.00}&\textbf{95.24}\\
         
         \text{white\_pocky} & 0.265 & 89.55 &  94.27&42.83&54.65& 17.19 & 47.45 & 24.93& 26.71&\textbf{90.83} &\textbf{94.70} \\
    \midrule
         \text{Average}  & 0.239 & 69.67  &85.88 &49.02&62.44& 13.25 & 37.38& 46.86& 55.74& \textbf{73.99}&\textbf{88.10} \\
    \bottomrule
    \end{tabular}}
\end{table*}

\begin{figure*}
\centering
\includegraphics[width=1.\linewidth]{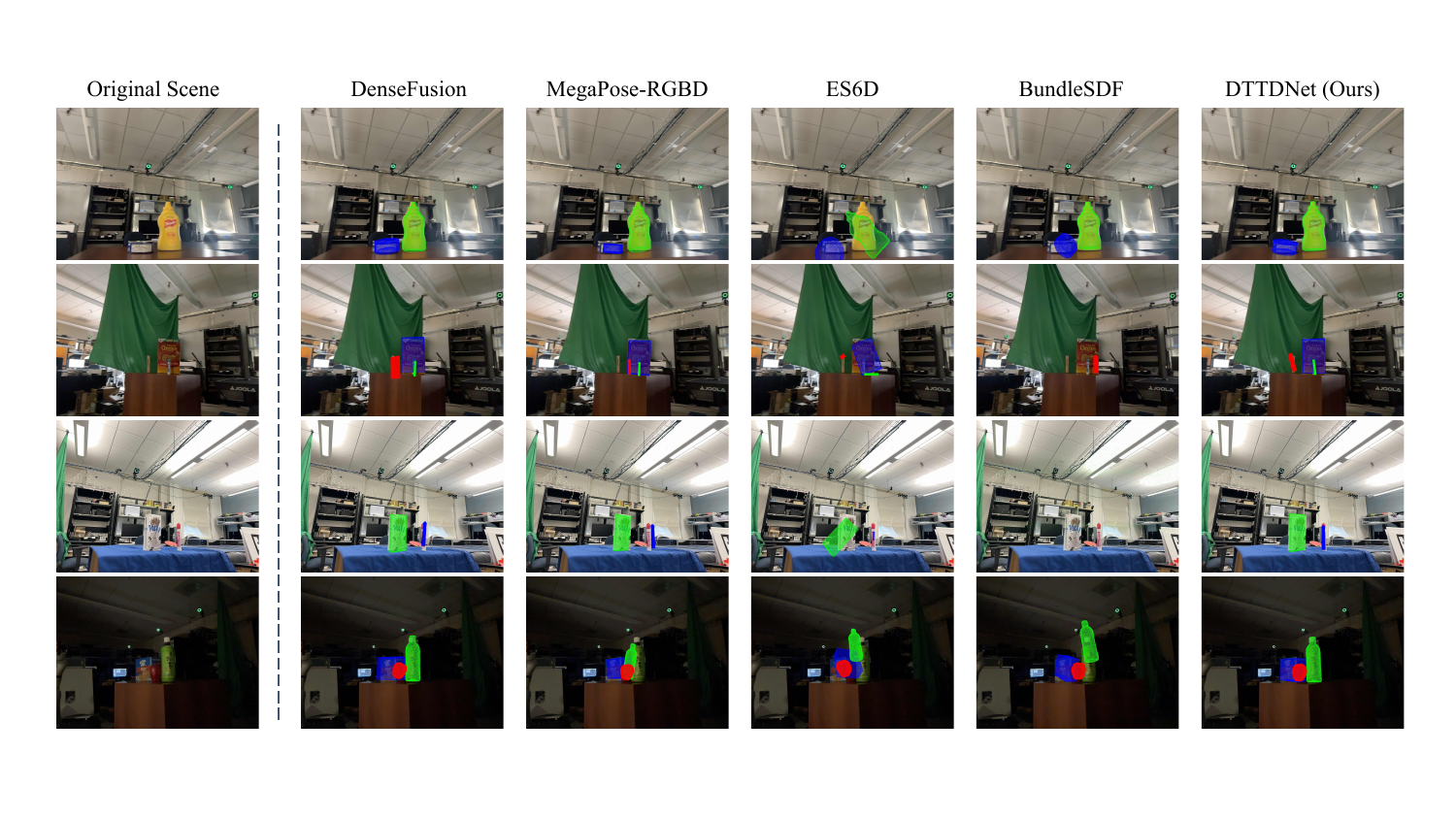}
\vspace{-48pt}
\caption{\textbf{Qualitative evaluation of different methods.} To further validate our approach, we provide visual evidence of our model's effectiveness in challenging occlusion scenarios and varying lighting conditions, where other models' predictions fail but ours remain reliable. It should be noted that BundleSDF \cite{wen2023bundlesdf} fails to reconstruct 3D objects in some scenes, resulting in the absence of annotations for such objects.}
\label{fig:qual_results}
\end{figure*}

\subsection{Evaluation Metrics}
We evaluate baselines with the average distance metrics ADD and ADD-S according to previous protocols \cite{xiang2018posecnn, DTTD}. Suppose $R$ and $t$ are ground truth rotation and translation and $\tilde{R}$ and $\tilde{t}$ are the predicted counterparts. The ADD metric computes the mean of the pairwise distances between the 3D model points using ground truth pose $(R, t)$ and predicted pose $(\tilde{R}, \tilde{t})$:
\begin{equation}
    \mathrm{ADD} = \frac{1}{m}\sum_{x \in M} \| (Rx + t) - (\tilde{R}x + \tilde{t})\|,
\end{equation}
where $M$ denotes the point set sampled from the object's 3D model and $x$ denotes the point sampled from $M$. 

The ADD-S metric is designed for symmetric objects when the matching between points could be ambiguous:
\begin{equation}
    \mathrm{ADD{-}S} = \frac{1}{m}\sum_{x_1 \in M}\min_{x_2 \in M} \| (Rx_1 + t) - (\tilde{R}x_2 + \tilde{t})\|.
\end{equation}

Following previous protocols \cite{xiang2018posecnn, wang2019densefusion, liu2021stereobj1m, liu2023gen6d, labbé2022megapose}, a 3D pose estimation is deemed accurate if the average distance error falls below a predefined threshold. Two widely-used metrics are employed in our work, namely ADD/ADD-S AUC and ADD/ADD-S(1cm). For commonly used ADD/ADD-S AUC, we calculate the Area Under the Curve (AUC) of the success-threshold curve over different distance thresholds, where the threshold values are normalized between 0 and 1. On the other hand, ADD/ADD-S(1cm) is defined as the percentage of pose error smaller than the 1cm threshold.

\begin{table*}[h]
    \centering
    \caption{\textbf{Comparison between 2 datasets with diverse 6DoF pose estimation baselines.} We evaluate the results as the ADD-S AUC on the overlapping objects between the YCB video dataset and the DTTD-Mobile (DTTD-M) dataset, higher is better. }
    \label{tab:eval_dttb_ycb}
    \resizebox{1.0 \textwidth}{!}{
    \begin{tabular}{|l|c|c|c|c|c|c|c|c|c|c|c|c|}
    \toprule
        & \multicolumn{2}{|c|}{\emph{depth-ADD}}
        & \multicolumn{2}{|c|}{DenseFusion \cite{wang2019densefusion}} 
         & \multicolumn{2}{|c|}{MegaPose-RGBD \cite{labbé2022megapose}}  
         & \multicolumn{2}{|c|}{ES6D \cite{mo2022es6d}} 
         & \multicolumn{2}{|c|}{BundleSDF 
 \cite{wen2023bundlesdf}} 
         & \multicolumn{2}{|c|}{DTTDNet (Ours)}
\\
    \midrule
         Object & 
         \multicolumn{1}{|c|}{YCB} & \multicolumn{1}{|c|}{DTTD-M} &
         \multicolumn{1}{|c|}{YCB} &
         \multicolumn{1}{|c|}{DTTD-M} &
         \multicolumn{1}{|c|}{YCB} &
         \multicolumn{1}{|c|}{DTTD-M} &
         \multicolumn{1}{|c|}{YCB}  &
         \multicolumn{1}{|c|}{DTTD-M}  & 
         \multicolumn{1}{|c|}{YCB}  &
         \multicolumn{1}{|c|}{DTTD-M}  &
         \multicolumn{1}{|c|}{YCB}  &
         \multicolumn{1}{|c|}{DTTD-M} 
\\
    \midrule
         
         \text{tomato\_can} & 0.011 & 0.222 & 93.70 &  93.42& 86.11 & 84.48 & 89.02 & 56.17 & 68.27 &  93.65& \textbf{96.69} & \textbf{94.01} \\
         
         \text{mustard\_bottle} & 0.005 & 0.184 & 95.90 & 91.31 & 87.41 & 85.38 & 93.13 & 52.56 & \textbf{98.21} & \textbf{92.99} & 97.39 & 92.15\\
         
         \text{tuna\_can} & 0.013 & 0.278 & 94.90 &  79.94 & 91.03 & 22.11 & 74.86 & 26.86 & 91.11 & 37.94 & \textbf{95.78} & \textbf{87.05}\\
    \midrule

         \text{average} & 0.009 & 0.228 & 94.83 &  88.22 & 88.18 & 63.99 & 85.67 & 45.20 & 85.86 & 74.86 & \textbf{96.62} & \textbf{91.07}\\

    \bottomrule
    \end{tabular}}
\vspace{-12pt}
\end{table*}

\subsection{Experimental Results} 
In this section, we compare the performance of our method DTTDNet with four other 6DoF pose estimators, namely, BundleSDF \cite{wen2023bundlesdf}, MegaPose \cite{labbé2022megapose},  ES6D \cite{mo2022es6d}, DenseFusion \cite{wang2019densefusion}. While all four methods leverage the benefits of multimodal data from both RGB and depth sources, they differ in the extent to which they emphasize the depth data processing module. Quantitative experimental results are shown in Table \ref{tab:eval_dttb}. Qualitative examples are shown in Fig. \ref{fig:qual_results}.

BundleSDF \cite{wen2023bundlesdf} learns multi-view consistent shape and appearance of a 3D object using an object-centric neural signed distance field, leverages frame poses captured in-flight. However, the approach struggles with 3D object reconstruction when faced with large distances, low resolution, or insufficient frame sequence per viewpoint. BundleSDF \cite{wen2023bundlesdf} achieves an ADD AUC of 46.86 and an ADD-S AUC of 55.74. This method failed to reconstruct 8 out of 26 object-scene combinations in DTTD-Mobile. As shown in \ref{fig:qual_results}, the failure in 3D object reconstruction results in the absence of object pose tracking in the corresponding scenes. The BundleSDF, as shown in Figure \ref{fig:hero_img}, excludes objects in scenes lacking associated pose tracking. 

MegaPose \cite{labbé2022megapose} employs a coarse-to-fine process for pose estimation. The initial "coarse" module leverages both RGB and depth data to identify the most probable pose hypothesis. Subsequently, a more precise pose inference is achieved through the "render-and-compare" technique. Disregarding the noise in the depth data can also impair the effectiveness of their coarse module, consequently leading to failure in their refinement process. Even with the assistance of a refiner, MegaPose-RGBD \cite{labbé2022megapose} only manages to attain an ADD AUC of 49.02 and an ADD-S AUC of 62.44. Its damage and susceptibility to depth noise falls somewhere between DenseFusion \cite{wang2019densefusion} and ES6D \cite{mo2022es6d}.

DenseFusion \cite{wang2019densefusion} treats both modalities equally and lacks a specific design for the depth module, whereas ES6D \cite{mo2022es6d} heavily relies on depth data during training, using grouped primitives to prevent point-pair mismatch. However, due to potential interpolation errors in the depth data, this additional supervision can introduce erroneous signals to the estimator, resulting in inferior performance compared to DenseFusion \cite{wang2019densefusion}. DenseFusion \cite{wang2019densefusion} achieves 69.67 ADD AUC and 85.88 ADD-S AUC, whereas ES6D \cite{mo2022es6d} only achieves 13.25 ADD AUC and 37.38 ADD-S AUC.

In contrast, our approach harnesses the strengths of both RGB and depth modalities while explicitly designing robust depth feature extraction and selection. In comparison with the above baselines, our method achieves 73.31 ADD AUC and 87.82 ADD-S AUC, surpassing the state of the art with improvements of 1.94 and 3.64 percent in terms of ADD AUC and ADD-S AUC, respectively.

\begin{figure}[t]
        \centering
\includegraphics[width=1.0\linewidth]{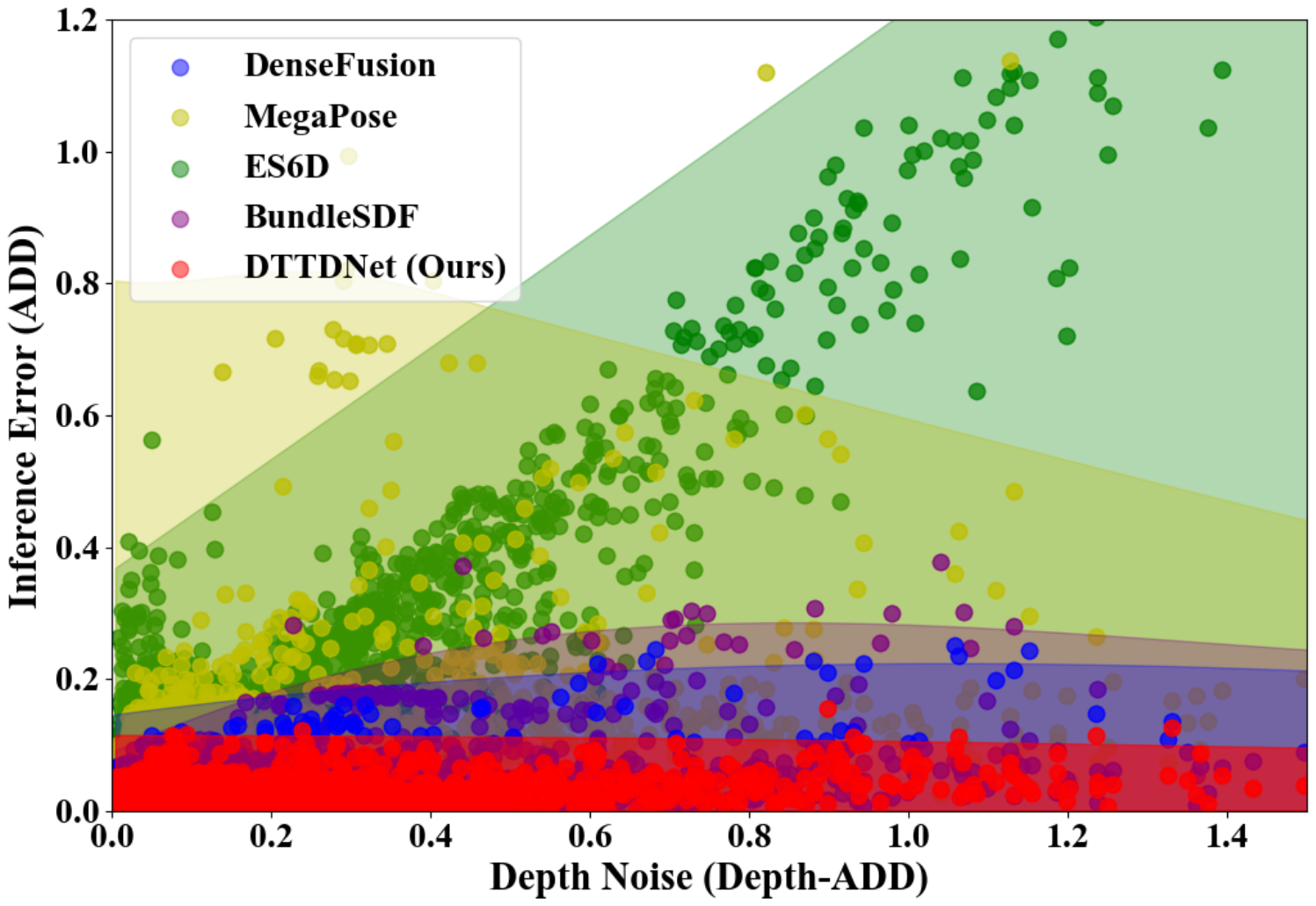}
    \caption{Shadow plot of the relation between the \textbf{depth noise} (\emph{depth-ADD}) and the \textbf{inference error} (ADD) of considered state-of-the-art methods and proposed DTTDNet.} 
\label{fig:hero_img_2}
\end{figure}

To assess the real-time applicability of DTTDNet, we benchmarked its inference speed on a single NVIDIA RTX A6000 GPU. Our model achieves an average inference time of 0.0172 seconds per object and 0.0378 seconds per frame, corresponding to 58.01 objects per second and 26.43 FPS. These results demonstrate that DTTDNet is well-suited for real-time 6DoF pose estimation, achieving efficient performance without sacrificing robustness or accuracy.

\subsection{Ablation Studies}
\label{ablation_studies}
In this section, we further delve into a detailed analysis of our own model, highlighting the utility of our depth robustifying module in handling challenging scenarios with significant LiDAR noise. 

\noindent \textbf{Evaluation of DTTDNet on other datasets.} To show that our proposed pose estimator could also be generalized to other domains, we evaluate our method on the YCB-Video dataset \cite{xiang2018posecnn}, which outperforms MegaPose \cite{labbé2022megapose} and ES6D \cite{mo2022es6d} by 6.87 and 7.90 points on ADD(S), respectively, when performing fair comparison (\textit{i.e.}, without any data pre-cleaning and iterative refinement). 

We share 3 overlapping objects between the YCB video dataset and our DTTD-Mobile dataset, and demonstrate the models' performance on these objects to examine performance drop of each baseline when shifting the datasets, as shown in Table \ref{tab:eval_dttb_ycb}. DTTDNet achieves the highest performance on both datasets, and shows less performance drop when occurring iPhone LiDAR noise. Detailed per-object evaluation is appended in the appendix.

\noindent \textbf{Robustness to LiDAR Depth Error.} To answer the question of whether our method exhibits robustness in the presence of significant LiDAR sensor noise when compared to other approaches, we further assess the \emph{depth-ADD} metric, as discussed in Section \ref{depthADD}, on DTTDNet versus the four baseline algorithms. Fig. \ref{fig:hero_img_2} illustrates the correlation between the model performance (ADD) of four methods and the quality of depth information (\emph{depth-ADD}) across various scenes, frames, and 1239 pose prediction outcomes for the 18 objects. Our approach ensures a stable pose prediction performance, even when the depth quality deteriorates, maintaining consistently low levels of ADD error overall. 


\begin{figure}[tb]
  \centering
    \includegraphics[width=0.9\linewidth]{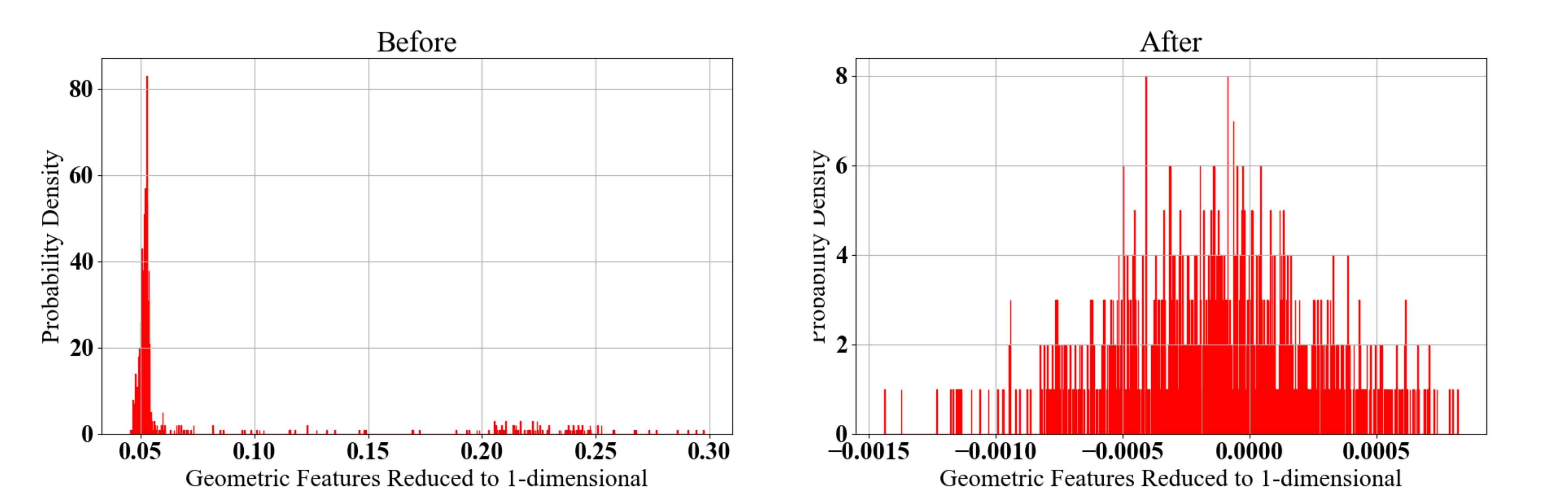}
    \caption{\textbf{Probability Distribution of Reduced Geometric Features.} \textit{Left}: before the GFF module. \textit{Right}: after the GFF module.} 
  \label{fig:hist_gff}
\end{figure}

\begin{table}[t]
    \centering
    \caption{\textbf{Effect of Depth Feature Filtering.} M8P4 denotes our model with a fusion stage consisting of 8-layer modality fusion and 4-layer point-wise fusion modules. This table shows the improvement of M8P4 with further incorporation of geometric feature filtering (GFF). }
    \label{tab:feature_robustifying}
    \resizebox{0.48 \textwidth}{!}{
    \begin{tabular}{l|cc|cc}
    \toprule
         Methods  & \textbf{ADD AUC}& ADD-S AUC  &  ADD (1cm) & ADD-S (1cm) \\
    \midrule
          M8P4& 72.03&86.44 &19.86&\textbf{70.50} \\
          + GFF&\textbf{73.31}&\textbf{87.82}&\textbf{24.35}&66.16\\
    \bottomrule
    \end{tabular}}
\end{table}

\begin{table}[t]
    \caption{\textbf{Effect of Object Geometry Augmented CDL.} This table depicts the enhancement in model performance when switching the reference point set from being reliant on the depth map to being augmented by the object model.}
    \label{tab:CDL}
    \centering
    \resizebox{0.48 \textwidth}{!}{
    \begin{tabular}{l|c|cc|cc}
    \toprule
         Methods & CDL supervised by & \textbf{ADD AUC}& ADD-S AUC & ADD (1cm) & ADD-S (1cm) \\
    \midrule
          \multirow{2}{*}{M8P4+GFF}& LiDAR depth &73.31&87.82&24.35&66.16\\
          & CAD model& \textbf{73.99}&\textbf{88.10}  & \textbf{25.85}& \textbf{67.75} \\
    \bottomrule
    \end{tabular}}
\vspace{-12pt}
\end{table}

\begin{table}[t]
    \centering
    \caption{\textbf{Effect of Layer Number in 2 Fusion Stages.} It shows DTTDNet with different layer number combinations in the fusion stages with one $\circ$ denoting one layer. For all combinations, CDL is used in the geometric feature extraction stage. }
    \label{tab_fusion_combo}
    \resizebox{0.48 \textwidth}{!}{
    \begin{tabular}{cc|cc|cc}
    \toprule
    \multicolumn{2}{c|}{Layer Num of \#} & \multicolumn{4}{|c}{Metrics} \\
    \midrule
          Modality Fusion&Point-wise Fusion &  \textbf{ADD AUC} &ADD-S AUC  & ADD (1cm) & ADD-S (1cm)   \\
    \midrule
         $\circ\circ$ &$\circ$  & 70.73& 85.42& 22.91& 67.75\\
         $\circ\circ$&$\circ\circ$&71.37&86.69&15.74&64.44\\
    $\circ\circ$& $\circ\circ$$\circ\circ$ & 72.06& 86.37 & 19.57&68.37 \\
    \midrule
    $\circ\circ$& $\circ$ & 70.73& 85.42 & 22.91& 67.75\\
         $\circ\circ$$\circ\circ$&$\circ\circ$  & 71.76& 88.23&20.01&69.03\\
         $\circ\circ$$\circ\circ$$\circ\circ$$\circ\circ$&$\circ\circ$$\circ\circ$ & 72.03& 86.44&19.86&70.50\\
    \bottomrule
    \end{tabular}}
\vspace{-12pt}
\end{table}

\noindent \textbf{Effect of Depth Feature Filtering module.} 
Table \ref{tab:feature_robustifying} illustrates the improvement in ADD AUC metrics achieved by our method when integrating geometric feature filtering module. To provide a detailed insight into the impact of the GFF module, we conducted principal component analysis (PCA) on both the initial geometric tokens encoded by the PointNet and the filtered version after applying the GFF module, i.e., projected the embedding to a 1-D array with its dominant factor. 
We visualize the geometric embedding both before and after the application of the GFF module by generating histograms of the dimensionally reduced geometric tokens, as shown in Fig. \ref{fig:hist_gff}. 
The distribution of these tokens, as shown in the right subplot, becomes more balanced and uniform after learning-based filtering through the GFF module. The enhanced ADD AUC performance can be attributed to the balanced distribution achieved through the use of the depth robustifying module.

\noindent \textbf{Effect of Geometry Augmented CDL.} We replaced the reference point set for CDL with measured LiDAR depth data as a baseline to demonstrate the effectiveness of our design choice when supervised by the point cloud sampled from 3D object models. In Table \ref{tab:CDL}, we conduct a performance comparison of our approach with these two reference point choices, 
our design choice achieved higher ADD AUC and ADD-S AUC, as well as higher performance in the more stringent metric, ADD/ADD-S 1cm (Table \ref{tab:CDL}).


\noindent \textbf{Effect of Layer Number Variation in Fusion Stages.} Table \ref{tab_fusion_combo} display the variations brought about by increasing the number of layers at different fusion stages. Overall, adding layer number increases the model's performance in terms of ADD AUC. As we proportionally increase the total number of layers in the modality fusion or point-wise fusion, sustained improvement is observed. 


\section{Conclusion}
We have presented DTTDNet as a novel digital-twin localization algorithm to bridge the performance gap for 3D object tracking in mobile enviroments and with critical requirements of accuracy. At the algorithm level, DTTDNet is a transformer-based 6DoF pose estimator, specifically designed to navigate the complexities introduced by noisy depth data. At the experiment level, we introduced a new RGBD dataset captured using iPhone 14 Pro, expanding our approach to iPhone sensor data. Through extensive experiments and ablation analysis, we have examined the effectiveness of our method in being robust to erroneous depth data. Additionally, our research has brought to light new complexities associated with object tracking in dynamic AR environments. 

\section{Acknowledgment}
This work is supported by the FHL Vive Center at the University of California, Berkeley. We gratefully acknowledge our other team members: Tianjian Xu and Kathy Zhuang for their support with the iPhone capturing app and BundleSDF evaluation on our dataset; Xiang Zhang for his assistance in the further robotic dataset extension\footnote{DTTD-Fanuc is an annotated dataset specifically designed for robotic grasping and manipulation with cheaper and low-resolution sensors. It is publicly available in our DTTDNet GitHub repository.}; Weiyu Feng and Chenfeng Xu for their valuable discussions and academic insights during the development of this project.



{
    \small
    \bibliographystyle{ieeenat_fullname}
    \bibliography{main}
}

\clearpage
\setcounter{page}{1}
\maketitlesupplementary


\section{Related Work}

\noindent \textbf{6DoF Pose Estimation Algorithms.}
The majority of data-driven approaches for object pose estimation revolve around utilizing either RGB images \cite{labbe2020, peng2019pvnet, xiang2018posecnn, zakharov2019} or RGBD images \cite{He_2021_CVPR, He_2020_CVPR, wang2019densefusion, mo2022es6d, Jiang_2022_CVPR} as their input source. RGBD remains mainstream in industrial environments requiring higher precision. However, due to the high cost of accurate depth sensors, finding more robust solutions compatible with inexpensive and widely used sensors is a problem we aim to address. 

Methods \cite{He_2021_CVPR, He_2020_CVPR, wang2019densefusion, mo2022es6d, Jiang_2022_CVPR} that relied on depth maps advocated for the modality fusion of depth and RGB data to enhance inference capabilities. To effectively fuse multi-modalities, Wang et al. \cite{wang2019densefusion} introduced a network architecture capable of extracting and integrating dense feature embedding from both RGB and depth sources. Due to its simplicity, this method achieved high efficiency in predicting object poses. In more recent works \cite{He_2020_CVPR, He_2021_CVPR, He_2022_CVPR}, performance improvements were achieved through more sophisticated network architectures. For instance, He et al. \cite{He_2021_CVPR} proposed an enhanced bidirectional fusion network for key-point matching, resulting in high accuracy on benchmarks such as YCB-Video \cite{xiang2018posecnn} and LINEMOD \cite{10.1007/978-3-642-37331-2_42}. However, these methods exhibited reduced efficiency due to the complex hybrid network structures and processing stages. Addressing symmetric objects, Mo et al. \cite{mo2022es6d} proposed a symmetry-invariant pose distance metric to mitigate issues related to local minima. On the other hand, Jiang et al. \cite{Jiang_2022_CVPR} proposed an L1-regularization loss named abc loss, which enhanced pose estimation accuracy for non-symmetric objects.

Besides the RGBD approach, studies following the RGB-only approach often rely on incorporating additional prior information and inductive biases during the inference process. These requirements impose additional constraints on the application of 3D object tracking on mobile devices. Their inference process can involve utilizing more viewpoints for similarity matching \cite{liu2023gen6d, labbé2022megapose} or geometry reconstruction \cite{sun2022onepose, wen2023foundationpose}, employing rendering techniques \cite{labbé2022megapose, nguyen2022templates, Cai_2022_CVPR} based on precise 3D model or leveraging an additional database for viewpoint encoding retrieval \cite{Cai_2022_CVPR}. During the training phase, these approaches typically draw upon more extensive datasets, such as synthetic datasets, to facilitate effective generalization within open-set scenarios. However, when confronted with a limited set of data samples, their performance does not surpass that of closed-set algorithms in cases where there is a surplus of prior information available and depth map loss.

\begin{figure*}[h]
        \centering
        \setlength{\abovecaptionskip}{0.cm}
        \includegraphics[width=1.\linewidth]{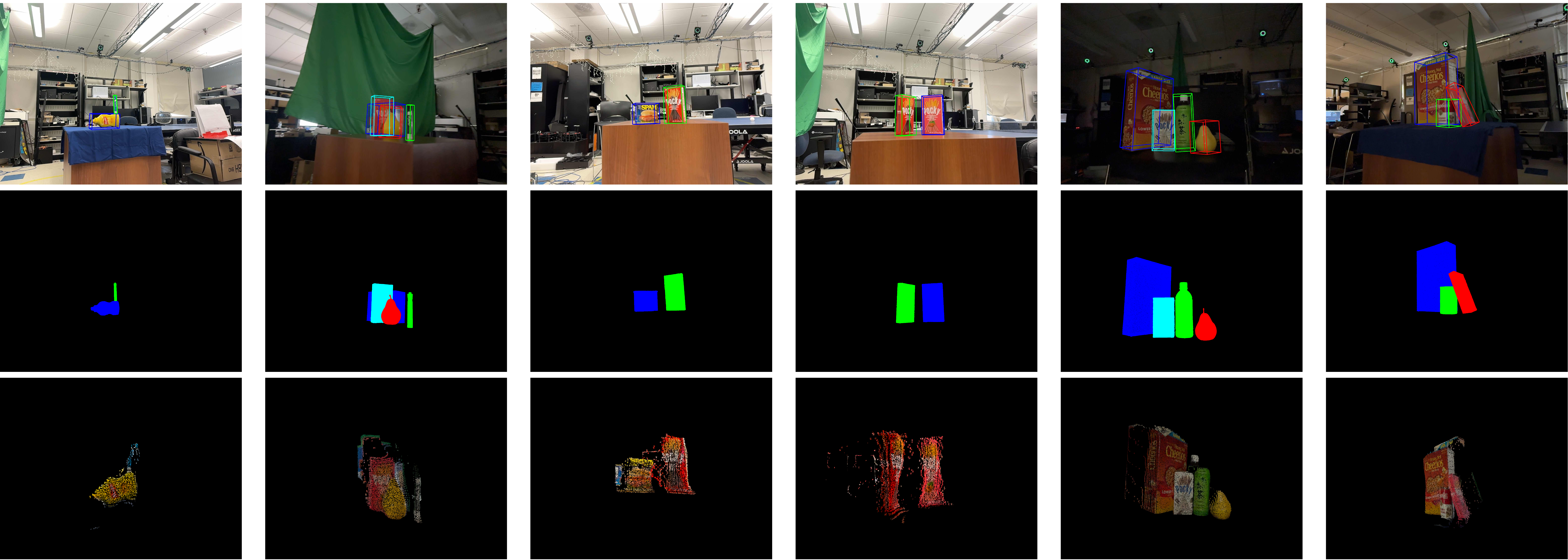}
    \caption{\textbf{Sample visualizations of our dataset.} \textit{First row:} Annotations for 3D bounding boxes. \textit{Second row:} Corresponding semantic segmentation labels. \textit{Third row:} Zoomed-in LiDAR depth visualizations.}
    \label{fig:scenes}
\end{figure*}

\noindent \textbf{3D Object Tracking Datasets}
Existing object pose estimation algorithms are predominantly tested on a limited set of real-world 3D object tracking datasets \cite{10.1007/978-3-642-37331-2_42, marion2018label, xiang2018posecnn, liu2020keypose, hodan2017tless, liu2021stereobj1m, DTTD, ycb1, ycb2, ycb3}, which often employ depth-from-stereo sensors or time-of-flight (ToF) sensors for data collection. Datasets like YCB-Video \cite{xiang2018posecnn}, LINEMOD \cite{10.1007/978-3-642-37331-2_42}, StereoOBJ-1M \cite{liu2021stereobj1m}, and TOD \cite{liu2020keypose} utilize depth-from-stereo sensors, while TLess \cite{hodan2017tless} and DTTD \cite{DTTD} deploy ToF sensors, specifically the Microsoft Azure Kinect, to capture meter-scale RGBD data. However, the use of cameras with depth-from-stereo sensors may not be an optimal platform for deploying AR software, because stereo sensors may degrade rapidly at longer distances \cite{Haggag2013MeasuringDA} and may encounter issues with holes in the depth map when stereo matching fails. In our pursuit of addressing the limitations of existing datasets and ensuring a more realistic dataset captured with mobile devices, we opt to collect RGBD data using the iPhone 14 Pro.

\noindent \textbf{iPhone-based Datasets for 3D Applications.}
Several datasets utilize the iPhone as their data collection device for 3D applications, such as ARKitScenes \cite{baruch2021arkitscenes}, MobileBrick \cite{li2023mobilebrick}, ARKitTrack \cite{zhao2023arkittrack}, and RGBD Dataset \cite{he2021towards}. These datasets were constructed to target applications from 3D indoor scene reconstruction, 3D ground-truth annotation, depth-map pairing from different sensors, to RGBD tracking in both static and dynamic scenes. However, most of these datasets did not specifically target the task of 6DoF object pose estimation. Our dataset provides a distinct focus on this task, offering per-pixel segmentation and pose labels. This enables researchers to delve into the 3D localization tasks of objects with a dataset specifically designed for this purpose. The most relevant work is from OnePose \cite{sun2022onepose}, which is an RGBD 3D dataset collected by iPhone. However, their dataset did not provide 3D models for close-set settings, and they utilized automatic localization provided by ARKit for pose annotation, which involved non-trivial error for high-accuracy 6DoF pose estimation. On the other hand, we achieve higher localization accuracy with the OptiTrack professional motion capture system to track the iPhone camera’s real-time positions as it moves in 3D.

\section{More Dataset Description}
\subsection{Data Acquisition}
Apple's ARKit framework\footnote{https://developer.apple.com/documentation/arkit/} enables us to capture RGB images from the iPhone camera and scene depth information from the LiDAR scanner synchronously. We leverage ARKit APIs to retrieve $1920 \times 1440$ RGB images and $256 \times 192$ depth maps at a capturing rate of 30 frames per second. Despite the resolution difference, both captured RGB images and depth maps match up in the aspect ratio and describe the same scene. Alongside each captured frame, DTTD-Mobile stores the camera intrinsic matrix and lens distortion coefficients, and also stores a 2D confidence map describing how the iPhone depth sensor is confident about the captured depth at the pixel level. In practice, we disabled the auto-focus functionality of the iPhone camera during data collection to avoid drastic changes in the camera's intrinsics between frames, and we resized the depth map to the RGB resolution using nearest neighbor interpolation to avoid depth map artifacts. 

To track the iPhone's 6DoF movement, we did not use the iPhone's own world tracking SDK. Instead, we follow the same procedure as in \cite{DTTD} and use the professional OptiTrack motion capture system for higher accuracy. For label generation, we also use the open-sourced data annotation pipeline provided by \cite{DTTD} to annotate and refine ground-truth poses for objects in the scenes along with per-pixel semantic segmentation. Some visualizations of data samples are illustrated in Fig. \ref{fig:scenes}. Notice that the scenes cover various real-world occlusion and lighting conditions with high-quality annotations. Following previous dataset protocols \cite{DTTD, xiang2018posecnn}, we also provide synthetic data for scene augmentations used for training. 

The dataset also provides 3D models of the 18 objects as illustrated in the main paper. These models are reconstructed using the iOS Polycam app via access to the iPhone camera and LiDAR sensors. To enhance the models, Blender \footnote{https://www.blender.org/} is employed to repair surface holes and correct inaccurately scanned texture pixels. 

\subsection{Train/Test Split} 
DTTD-Mobile offers a suggested train/test partition as follows. The training set contains 8622 keyframes extracted from 88 video sequences, while the testing set contains 1239 keyframes from 12 video sequences. To ensure a representative distribution of scenes with occluded objects and varying lighting conditions, we randomly allocate them across both the training and testing sets. Furthermore, for training purposes of scene augmentations, we provide 20,000 synthetic images by randomly placing objects in scenes using the data synthesizer provided in \cite{DTTD}.

\begin{figure*} 
        \centering
\includegraphics[width=1.0\linewidth]{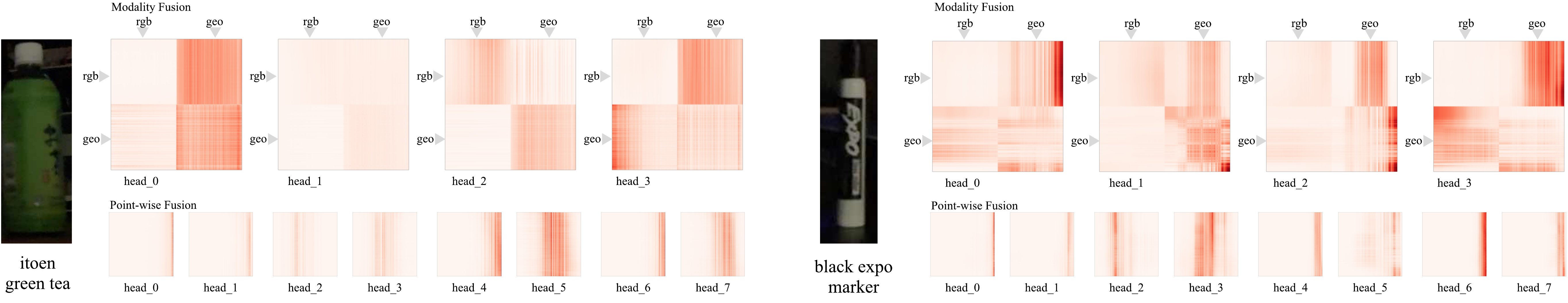}
    \caption{Examples of attention map output visualize of both modality fusion stage (the larger maps in the first row) and point-wise fusion stage (the smaller ones in the second row) on two objects (\textit{itogen\_green\_tea} and \textit{black\_marker}). Due to the different ways we concentrate features in the two fusion stages, the token sequence length in modality fusion is twice that in the point-wise fusion process. For the attention maps produced in the final layer of modality fusion and point-wise fusion, they are of sizes $2000\times2000$ and $1000\times 1000$, respectively.}
\label{fig:attention}
\end{figure*}

\section{More Implementation Details}
\subsection{Details on RGBD Feature Fusion}
\noindent \textbf{Attention Mechanism.}\label{attention} For both modality fusion and point-wise fusion stage, the scaled dot-product attention is utilized in the self-attention layers:
\begin{equation}
\mathrm{Attention}(\mathbf{Q}, \mathbf{K}, \mathbf{V})_i = \sum_j \dfrac{\exp(\mathbf{q}_i^T \mathbf{k}_j / \sqrt{d_\text{head}})}{\sum_k \exp(\mathbf{q}_i^T \mathbf{k}_k / \sqrt{d_\text{head}})} \mathbf{v}_j,
\end{equation}
where query, key, value, and similarity score are denoted as $q$, $k$, $v$, and $s$. The distinction between two fusion stages lies in the token preparation prior to the linear projection layer. It results in varying information contained within the query, key, and value. 

The key idea in the first fusion stage is to perform local per-point fusion in a cross-modality manner so that we can make predictions based on each fused feature. Each key or query carries only one type of modal information before fusion, allowing different modalities to equally interact with each other through dot-product operations. It exerts a stronger influence when the RGB and geometric representations produce higher similarity. 

In the second stage, where we integrate two original single-modal features with the first-stage feature into each point, we calculate similarities solely among different points. The key idea is to enforce attention layers to further capture potential relationships among multiple local features. A skip connection is employed in a concentrating manner between two fusion outputs so that we can make predictions based on per-point features generated in both the first and second stages.

\noindent \textbf{Modality Fusion.} 
The objective of this module is to combine geometric embedding $g$ and RGB embedding $c$ produced by single-modal encoders in a cross-modal fashion. Drawing inspiration from {ViLP} \cite{kim2021vilt}, both types of embedding are linearly transformed into a token sequence ($\in\mathbb{R}^{N\times d_{emb}}$). Before entering the modality fusion module $E_1$, these features are combined along the sequence length direction, i.e., all feature embedding is concentrated into a single combined sequence, where the dimension remains $d_{emb}$, and the sequence length becomes twice the original length.  
\begin{equation}
f_1=E_{1}\left[c\oplus g\right]\in\mathbb{R}^{d_{f_1}\times 2N}
\end{equation}

where the operation symbol "$\oplus$" denotes concentrating along the row direction. It is then reshaped into the sequence $f_1'$ with the length of N and dimension of $2d_{f_1}$ in order to adapt the point-wise transformer encoder in the next fusion stage. This step enables the model's attention mechanism to effectively perform cross-modal fusion tasks.

\noindent \textbf{Point-Wise Fusion.} 
The goal of this stage is to enhance the integration of information among various points. The primary advantage of our method over the previous work \cite{He_2021_CVPR} is that our model can calculate similarity scores not only with the nearest point but also with all other points, allowing for more comprehensive interactions. In order to enable the point-wise fusion to effectively capture the similarities between different points, we merge the original RGB token sequence $c$ and the geometric token sequence $g$ together with the output embedding sequence $m'$ from the modality fusion module along the feature dimension direction. The combined sequence input $\left[c^T\oplus g^T\oplus (f_1')^T\right]^T\in \mathbb{R}^{(2d_{emb}+2d_{f_1})\times N}$ is then fed into the point-wise transformer encoder $E_{2}$ to acquire the final fusion: 
\begin{equation}
f_2= E_{2}\left [c^T \oplus  g^T \oplus  (f_1')^T\right ]^T\in \mathbb{R}^{d_{f_2}\times N}
\end{equation}

\noindent \textbf{Attention Map Visualization.} To visualize what our fusion module learns during the training process, we draw on previous studies \cite{trockman2023mimetic,guo2023robustifying} and represent our attention map as $a_{i,j}$ described in section \ref{attention}.
Taking two objects (\textit{itoen\_green\_tea} and \textit{black\_marker}) as examples, Fig. \ref{fig:attention} displays the attention maps produced by different attention heads in the two fusion stages. We showcase the attention maps generated by the modality fusion and point-wise fusion at their respective final layers. The modality fusion part reveals distinct quadrant-like patterns, reflecting differences in how the two modalities fuse. The lower-left and upper-right quadrants offer insights into the degree of RGB and geometric feature fusion. The point-wise fusion part exhibits a striped pattern and shows that it attends to the significance of specific tokens during training.

\begin{table*}
    \centering
    \caption{\textbf{Comparison with diverse 6DoF pose estimation baselines on YCB video dataset.} We evaluate the results as the prior works \cite{wang2019densefusion} using ADD-S AUC and ADD-S (2cm) on all 21 objects, higher is better. Note that the left-most column indicates the per-object \emph{depth-ADD} error. Objects with names in bold are symmetric. }
    \label{tab:eval_dttb_ycb_final}
    \resizebox{1.0 \textwidth}{!}{
    \begin{tabular}{|l|c|c|c|c|c|c|c|c|c|c|c|}
    \toprule
        & \emph{depth-ADD}
        & \multicolumn{2}{|c|}{DenseFusion \cite{wang2019densefusion}} 
         & \multicolumn{2}{|c|}{MegaPose-RGBD \cite{labbé2022megapose}}  
         & \multicolumn{2}{|c|}{ES6D \cite{mo2022es6d}} 
         & \multicolumn{2}{|c|}{BundleSDF 
 \cite{wen2023bundlesdf}} 
         & \multicolumn{2}{|c|}{DTTDNet (Ours)}
\\
    \midrule
         Object & 
         Average &
         \multicolumn{1}{|c|}{ADD-S AUC} &
         \multicolumn{1}{|c|}{ADD-S (2cm)} &
         \multicolumn{1}{|c|}{ADD-S AUC} &
         \multicolumn{1}{|c|}{ADD-S (2cm)} &
         \multicolumn{1}{|c|}{ADD-S AUC}  &
         \multicolumn{1}{|c|}{ADD-S (2cm)}  & 
         \multicolumn{1}{|c|}{ADD-S AUC}  &
         \multicolumn{1}{|c|}{ADD-S (2cm)}  &
         \multicolumn{1}{|c|}{ADD-S AUC}  &
         \multicolumn{1}{|c|}{ADD-S (2cm)} 
\\
    \midrule
         \text{master\_chef\_can}& 0.005  & 95.20 & \textbf{100.00} & 79.11 & 69.88 & 82.47 & 73.56 & \textbf{97.05} & \textbf{100.00} & 96.32 & \textbf{100.00}\\
         
         \text{cracker\_box}  & 0.005 & 92.50 & \textbf{99.30} & 74.98 & 80.65 & 81.09 & 84.68 & 90.69 & 87.67 & \textbf{92.92} & 98.04\\
         
         \text{sugar\_box}& 0.009  & 95.10 & \textbf{100.00} &  81.42 & 90.95 & 95.97 & 97.80 & \textbf{97.79} & \textbf{100.00} & 96.76 & \textbf{100.00}\\
         
         \text{tomato\_can} & 0.011 & 93.70 & 96.90 & 86.11 & 94.79 & 89.02 & 92.71 & 68.27 & 70.00 & \textbf{96.69} & \textbf{99.17}  \\
         
         \text{mustard\_bottle} & 0.005 & 95.90 & \textbf{100.00} & 87.41 & 99.72 & 93.13 & 87.11 & \textbf{98.21} & \textbf{100.00} & 97.39 & \textbf{100.00}\\
         
         \text{tuna\_can} & 0.013 & 94.90 & \textbf{100.00}  & 91.03 & \textbf{100.00} & 74.86 & 74.22 & 91.11 & \textbf{100.00} & \textbf{95.78} & \textbf{100.00}\\
         
         \text{pudding\_box} & 0.005 & 94.70 & \textbf{100.00} & 89.65 & \textbf{100.00} & 90.13 & 98.60 & \textbf{97.67} & \textbf{100.00} & 93.24 & 95.33\\
         
         \text{gelatin\_box} & 0.011 & 95.80 & \textbf{100.00} & 87.17 & 99.07 & 97.39 & \textbf{100.00} & \textbf{98.46} & \textbf{100.00} & 97.97 & \textbf{100.00} \\
         
         \text{potted\_meat} & 0.008 & 90.10 & \textbf{93.10} & 77.88 & 80.81 & 78.56 & 75.46 & 62.00 & 58.22 & \textbf{93.56} & 92.04\\
         
         \text{banana} & 0.007  & 91.50 & 93.90 & 76.18 & 71.77 & 92.83 & 84.70 & \textbf{97.72} & \textbf{100.00} & 94.52 & \textbf{100.00}\\
         
         \text{pitcher\_base}  & 0.007 & 94.60 & \textbf{100.00} & 91.26 & \textbf{100.00} & 93.67 & 90.18 & \textbf{96.53} & \textbf{100.00} & 95.76 & \textbf{100.00}\\
         
         \text{bleach\_cleanser} & 0.007 & \textbf{94.30} & \textbf{99.80} & 82.97 & 74.05 & 88.12 & 87.76 & 69.67 & 71.72 & 93.30 & 99.61 \\
         
         \textbf{bowl} & 0.021  & 86.60 & 69.50 & 83.80 & 59.61 & 2.68 & 0.00 & \textbf{97.57} & \textbf{100.00} & 84.33 & 51.23\\
         
         \text{mug} & 0.010 & 95.50 & \textbf{100.00} & 86.63 & 97.64 &  88.58 & 89.15 & \textbf{97.09} & \textbf{100.00} & 97.00 & 99.53 \\
         
         \text{power\_drill} & 0.010 & 92.40 & 97.10 & 88.86 & 97.73 & 85.43 & 78.62 & \textbf{97.17} & \textbf{99.81} & 95.57 & \textbf{99.81} \\
         
         \textbf{wood\_block}& 0.007 & 85.50 & \textbf{93.40} & 35.55 & 0.41 & 29.56 & 1.24 & 19.57 & 0.00 & \textbf{87.63} & 90.50\\
         
         \text{scissors} & 0.010 &  \textbf{96.40} & \textbf{100.00} & 26.52 & 7.18 & 33.38 & 27.07 & 93.25 & 97.24 & 71.68 & 20.99 \\
         
         \text{large\_marker} & 0.011  & 94.70 & \textbf{99.20}  & 83.02 & 67.13 & 90.08 & 87.96 & 95.08 & 93.83 & \textbf{95.66} & 97.22\\
         
         \textbf{large\_clamp} & 0.011 & 71.60 & 78.50 & 85.93 & 90.03 & 43.74 & 17.84 & \textbf{96.77} & 99.16 & 90.99 & \textbf{99.44}\\

         \textbf{extra\_large\_clamp} & 0.012 & 69.00 & 69.50 &76.49& 88.32 & 66.77 & 69.35 & \textbf{95.15} & \textbf{100.00} & 89.70 & 93.11\\
         \textbf{foam\_brick} & 0.007 & 92.40 & \textbf{100.00} & 84.29 & 92.36 & 26.11 & 27.08 & 0.00 & 0.00 & \textbf{95.63} & \textbf{100.00}\\
    \midrule
         \text{Average}  & 0.009 & 91.20  & 95.30 & 82.64 & 84.81 & 78.14 & 75.35 & 86.31 & 87.64 & \textbf{94.19} & \textbf{96.14} \\
    \bottomrule
    \end{tabular}}
\end{table*}

\subsection{Hyperparameters}
\noindent \textbf{Details on Fusion Stages' Hyperparameters.} 
We extracted $1000$ of pixels from the decoded RGB representation corresponding to the same number of points in the LiDAR point set. Both extracted RGB and geometric features are linear projected to 256-D before fused together. In the final experiment results, we utilized an 8-layer transformer encoder with 4 attention heads for the modality fusion stage and a 4-layer transformer encoder with 8 attention heads for the point-wise fusion stage. 


\noindent \textbf{Training Strategies.} 
For our DTTDNet, learning rate warm-up schedule is used to ensure that our transformer-based model can overcome local minima in early stage and be more effectively trained. By empirical evaluation, in the first epoch, the learning rate $lr$ linearly increases from $0$ to $1e{-5}$. In the subsequent epochs, it is decreased using a cosine scheduler to the end learning rate $min\_lr=1e{-6}$. Additionally, following the approach of DenseFusion \cite{wang2019densefusion}, we also decay our learning rate by a certain ratio when the average error is below a certain threshold during the training process. Detailed code and parameters will be publicly available in our code repository. Moreover, we set the importance factor $\lambda$ of Chamfer distance loss to $0.3$ and the initial balancing weight $w$ to $0.015$ by empirical testing.

\section{More Experimental Details}

\subsection{Baseline Implementation Details}
For all the baseline methods \cite{wang2019densefusion, labbé2022megapose, mo2022es6d, wen2023bundlesdf} that we adopted, we did not integrate any additional iterative refinement processes (e.g., ICP \cite{besl1992method}) for fair comparison.

\noindent \textbf{ES6D \cite{mo2022es6d}.} 
We preprocessed the training datasets including both DTTD-Mobile and YCB video according to the original paper and official codebase of ES6D \cite{mo2022es6d}, including removing some high-noise data, normalizing 3D translation, averaging the xyz map, and filtering out outliers in the point cloud. For the test set, to ensure fair comparison with other baselines, we did not adopt certain noise reduction methods that require prior knowledge based on ground truth data, nor did we exclude some high-noise data samples. We ensured that all test samples were retained and had errors calculated as those in other baselines.

\noindent \textbf{BundleSDF \cite{wen2023bundlesdf}.} 
The object-centric camera pose coordinates outputted by BundleSDF \cite{wen2023bundlesdf} are based on its own embedded geometric reconstruction. In order to compute metrics in the same coordinate system and with the same CAD models as other baselines, we aligned the camera trajectory computed for each scene-object combination with the ground truth camera pose through trajectory-wise alignment based on the Umeyama algorithm \cite{umeyama1991least}.


\subsection{More Results on YCB video Dataset}
Due to the lower depth noise in the YCB video dataset (YCB's average \emph{depth-ADD} is 0.009, while DTTD-Mobile's average \emph{depth-ADD} is 0.239), we adopted a simplified DTTDNet model structure, omitting the GFF module, to accelerate training convergence. Additionally, the reference point set used for computing Chamfer distance loss was directly extracted from the depth map. Furthermore, regarding hyper-parameters, we chose 0 layers for modality fusion and 6 self-attention layers for point-wise fusion.

When examining the performance of each baseline on the YCB video dataset, as shown in \ref{tab:eval_dttb}. DTTDNet achieves the highest average performance, with an ADD-S AUC of 94.19 and an ADD-S (2cm) of 96.14. DenseFusion \cite{wang2019densefusion} is the second best, with an ADD-S AUC of 91.20 and an ADD-S (2cm) of 95.30. Although BundleSDF \cite{wen2023bundlesdf} shows strong performance across many object classes, it struggles with pose estimation for some objects, primarily due to its inability to reconstruct 3D models in the presence of occlusions. Its ADD-S AUC is 86.31, and its ADD-S (2cm) is 87.64. In contrast, MegaPose-RGBD \cite{labbé2022megapose} and ES6D \cite{mo2022es6d} lag behind in performance, particularly in the number of object classes in which they perform the best, with ADD-S AUC scores of 82.64 and 78.14, and ADD-S (2cm) scores of 84.81 and 75.35, respectively.



\end{document}